%% file: example_paper.tex
\documentclass{article}

\usepackage{microtype}
\usepackage{graphicx}
\usepackage{subfigure}
\usepackage{booktabs} 
\usepackage{hyperref}

\usepackage[accepted]{icml2024}

\usepackage{amsmath}
\usepackage{amssymb}
\usepackage{mathtools}
\usepackage{amsthm}

\usepackage[capitalize,noabbrev]{cleveref}

\theoremstyle{plain}
\newtheorem{theorem}{Theorem}[section]

\theoremstyle{definition}
\newtheorem{definition}[theorem]{Definition}

\theoremstyle{remark}

\usepackage[textsize=tiny]{todonotes}

\icmltitlerunning{Evolving Subnetwork Training for Large Language Models}

\begin{document}

\twocolumn[
\icmltitle{Evolving Subnetwork Training for Large Language Models}



\icmlsetsymbol{equal}{*}

\begin{icmlauthorlist}
\icmlauthor{Hanqi Li}{xlance}
\icmlauthor{Lu Chen}{xlance,suzhou}
\icmlauthor{Da Ma}{xlance}
\icmlauthor{Zijian Wu}{xlance}
\icmlauthor{Su Zhu}{aispeech}
\icmlauthor{Kai Yu}{xlance,suzhou}
\end{icmlauthorlist}

\icmlaffiliation{xlance}{X-LANCE Lab, Department of Computer Science and Engineering, MoE Key Lab of Artificial Intelligence, SJTU AI Institute, Shanghai Jiao Tong University, Shanghai, China}
\icmlaffiliation{suzhou}{Suzhou Laboratory, Suzhou, China}
\icmlaffiliation{aispeech}{AISpeech Co., Ltd., Suzhou, China}
\icmlcorrespondingauthor{Lu Chen}{chenlusz@sjtu.edu.cn}
\icmlcorrespondingauthor{Kai Yu}{kai.yu@sjtu.edu.cn}

\icmlkeywords{Machine Learning, ICML}

\vskip 0.3in
]



\printAffiliationsAndNotice{}  

\begin{abstract}
Large language models have ushered in a new era of artificial intelligence research. However, their substantial training costs hinder further development and widespread adoption. In this paper, inspired by the redundancy in the parameters of large language models, we propose a novel training paradigm: Evolving Subnetwork Training (EST). EST samples subnetworks from the layers of the large language model and from commonly used modules within each layer, Multi-Head Attention (MHA) and Multi-Layer Perceptron (MLP). By gradually increasing the size of the subnetworks during the training process, EST can save the cost of training. We apply  EST to train  GPT2 model and TinyLlama model, resulting in 26.7\% FLOPs saving for GPT2 and 25.0\% for TinyLlama without an increase in loss on the pre-training dataset. Moreover, EST leads to performance improvements in downstream tasks, indicating that it benefits generalization. Additionally, we provide intuitive theoretical studies based on training dynamics and Dropout theory to ensure the feasibility of EST. Our code is available
at \href{https://github.com/OpenDFM/EST}{https://github.com/OpenDFM/EST}.
\end{abstract}

\input{sections/introduction}

\input{sections/related_works}

\input{sections/method}

\input{sections/experiment}

\input{sections/theory}

\input{sections/conclusion}

\section*{Impact Statement}
This work's ethical impact is rooted in the ethical risks associated with large language models themselves. While there are numerous ethical risks linked to large language models, this paper primarily focuses on efficient training for such models, and thus, these ethical risks are not the main emphasis of this paper. Therefore, we believe it is not necessary to highlight them here.

The future societal consequences of this work primarily involve its impact on the environment and the applications of large language models. As this work helps reduce the training costs of large language models, it contributes to mitigating the carbon emissions caused by research and applications of such models, thereby aiding environmental conservation. Simultaneously, the cost savings may also facilitate a broader and more widespread application of large language models in society.

\section*{Acknowledgments}
This work is funded by the China NSFC Projects (92370206, U23B2057, 62106142 and 62120106006) and Shanghai Municipal Science and Technology Major Project (2021SHZDZX0102).


\nocite{langley00}

\input{example_paper.bbl}
\bibliography{example_paper}
\bibliographystyle{icml2024}

\input{sections/appendix}

\end{document}

%% file: sections/introduction.tex
\section{Introduction}
\label{Introduction}
Large language models (LLMs) have become significantly larger recently, bringing tremendous potential in Natural Language Processing~(NLP) tasks. The computational cost of training such large language models has become a bottleneck, hindering further development in research and applications. Additionally, the escalating hardware demands and increasing carbon footprints associated with training large language models are also crucial issues~\cite{schwartz2020green}. This highlights the importance of researching efficient algorithms for training large language models.

The enormous training cost of large language models stems from their massive number of parameters. For instance, the GPT3~\cite{gpt3} model has 175 billion parameters, requiring 355 GPU-years and incurring a training cost of \$4.6M.  However, numerous studies have highlighted the redundancy in the parameters of large language models, manifested in the over-parameterization~\cite{over_para} and conditional sparsity~\cite{conditional_sparsity} of these models. This inspires us to optimize the training process by exploring the possibility of not training the complete model at certain stages but focusing on training subnetworks, thereby reducing the overall training cost of large language models.

In this paper, we propose a novel training paradigm,  \textbf{Evolving  Subnetwork Training (EST)}, towards efficient training for large language models. EST paradigm consists of two main components: 1) sample from the large language model for subnetwork training. We maintain the complete model and sample subnetworks in each training step from the model across three dimensions, including the number of attention heads, the intermediate size of multi-layer perceptron, and the total number of Transformer layers. 2) We design a sampling scheduler to plan the training process. By increasing the size of subnetworks during training and, finally, training the complete model, EST achieves training acceleration. 

To demonstrate the effectiveness of EST, we first conduct experiments on GPT2 model~\cite{gpt2}, and conduct scale-up experiments on 1.1B TinyLlama~\cite{tinyllama} model. The results show that: 1) EST saves 26.7\%  training cost for GPT2 model and 25.0\% training cost for TinyLlama model, with a comparable loss on the pre-training dataset. 2) Models trained by EST achieve even higher downstream performance, indicating that EST benefits model generalization.

Furthermore, we dive into theoretical studies to answer the following two questions: 1) why EST can save training cost without compromising model performance? 2) Why EST can benefit model generalization? We provide a comprehensive theoretical framework based on deep learning dynamics and Dropout theory to ensure the superiority and feasibility of EST. 

In general, the contributions of this work include the following aspects:
\begin{itemize}
\item[$\bullet$] We propose a novel model training paradigm, EST, achieving higher optimization efficiency of training large language models.
\item[$\bullet$]We conduct experiments on GPT2 and TinyLlama. The results show that EST saves training cost without sacrificing model performance and benefits generalization.
\item[$\bullet$]We provide intuitive theoretical studies to ensure the feasibility of EST. 
\end{itemize}

%% file: sections/related_works.tex
\section{Related Work}
\label{Related_Work}
\subsection{Efficient Training for Large Language Models}
Many previous works aim at improving the efficiency of training large language models, ranging from addressing low-level hardware computations and memory bottlenecks to designing high-level training strategies. \\There are numerous approaches to overcome the computation bottleneck of Transformer-based models. FlashAttention~\cite{flashattention} identifies that the attention module is bottlenecked by memory access, and optimizes the process of attention computation, effectively reducing the training cost. Reformer~\cite{Reformer} approximates attention computation based on locality-sensitive hashing and  Performer~\cite{performer} simplifies attention computation with low-rank approximation.  \\Sparse training methods also benefit optimization efficiency. The main component of sparse training methods is the Mixture of Experts~(MoE). MoE methods~\cite{switch_transformer,glam} apply conditional computation according to different inputs in order to scale up models without significantly increasing training costs. The drawback of the MoE model is that its performance cannot match that of the dense model with an equivalent number of parameters. Another category of sparse training methods is based on the lottery ticket hypothesis~\cite{lottery,earlybert}, that certain subnetworks exhibit comparable performance to that of the original complete network. However, the sparsity generated by such methods is typically unstructured, making it challenging to translate into acceleration on general GPU devices. Monarch~\cite{dao2022monarch} and Pixelated Butterfly~\cite{dao2021pixelated} lower the training overhead of models without compromising their performance by introducing structured sparsity in matrix operations, which are more low-level and can be combined with our approach for complementary benefits. \citet{mada2024sparsityaccelerated} leverages sparsity in pre-trained LLMs to expedite the training process.

In this paper, we mainly focus on the design of a top-level training strategy, which is orthogonal to these approaches. Different from MoE methods that choose subnetworks based on input tokens,  our method samples subnetworks randomly.

\subsection{Incremental  Training}
\label{Incremental Training}
The most similar prior works are those called incremental training methods~\cite{shen2022staged}. This kind of work typically starts from smaller models and gradually scales up to larger ones. Incremental training methods are effective in both the NLP and CV domains since they reduce the time needed for model training and enhance training stability. Net2net~\cite{net2net} first reveals the feasibility of using the parameters of smaller models as initialization for larger model parameters and provides some operations for scaling up model sizes. \citet{shen2022staged} involves multi-stage training of the GPT2 model across both depth and width dimensions. bert2BERT~\cite{bert2bert} applies the principles of Net2net to pre-trained language models. MSG~\cite{MSG} employs a masking mechanism that sets newly expanded parameters to zero when scaling up small models. Unlike these methods that individually train smaller models, disregarding interactions between parameters, our approach EST maintains the complete model and samples subnetworks from it to train.

%% file: sections/method.tex
\section{Methodology}
\label{methodology}
In this section, we first review the most popular LLM architecture Transformer~\cite{vaswani2017attention} in Section~\ref{preliminaries}. In Section~\ref{subnetwork_training}, we discuss how to sample subnetworks from the Transformer model for subnetwork training. In Section~\ref{evolving_training}, we propose our efficient training paradigm, \textbf{E}volving \textbf{S}ubnetwork \textbf{T}raining~(EST).

\subsection{Preliminaries}
\label{preliminaries}

Recent large language models are mainly based on Transformer architecture. Before presenting our method, we first introduce the structure of Transformer, including two main components, multi-head attention~(MHA) and multi-layer perceptron~(MLP).

\textbf{Transformer Layer:}
 Let $\boldsymbol{\mathrm{X}}_{l-1}\in \mathbb{R}^{N\times d}$ denote the input sequence of layer $l$, where $N$ is the sequence length and $d$ denotes the hidden size. The sequence is processed iteratively by several Transformer layers with residual connection
\begin{equation}
\setlength{\abovedisplayskip}{8pt}
\setlength{\belowdisplayskip}{8pt}
\begin{aligned}
&\boldsymbol{\mathrm{X}}_l = \boldsymbol{\mathrm{X}}_{l-1} + \mathrm{Layer}_l(\boldsymbol{\mathrm{X}}_{l-1}), \forall{l\in\{1,2,...,N_L\}},
\end{aligned}
\end{equation}
where $N_L$ denotes the number of Transformer layers. Each Transformer layer is composed of one MHA module and one MLP module.

\textbf{MHA:}
MHA is used to mix information along the sequence axes to capture token-level dependencies. Let $N_H$ denote the number of heads and $d_k$ denote the dimension of each head. In layer $l$, for each head $i$, key, query, value projections are $\boldsymbol{\mathrm{W}}_{l,i}^Q$,$\boldsymbol{\mathrm{W}}_{l,i}^K$,$\boldsymbol{\mathrm{W}}_{l,i}^V\in \mathbb{R}^{d\times d_k}$ and the output projection is $\boldsymbol{\mathrm{W}}_{l,i}^O$. MHA is formulated as follows,
\begin{equation}
\begin{aligned}
&\boldsymbol{\mathrm{h}}_{l,i} = \mathrm{softmax}\left(\frac{\boldsymbol{\mathrm{X}}_{l-1}\boldsymbol{\mathrm{W}}_{l,i}^Q(\boldsymbol{ 
\mathrm{X}}_{l-1}\boldsymbol{\mathrm{W}}_{l,i}^K)^\mathsf{T}}{\sqrt{d_k}}\right)\boldsymbol{\mathrm{X}}_{l-1}\boldsymbol{\mathrm{W}}_{l,i}^V,\\
&\boldsymbol{\mathrm{X}}^{\mathsf{MHA}}_{l}= \sum_{i=1} ^{N_H} \boldsymbol{\mathrm{h}}_{l,i}\boldsymbol{\mathrm{W}}_{l,i}^O.
\end{aligned}
\end{equation}

\textbf{MLP:} 
MLP is used to  mixes information along the hidden dimension axes. It consists of two linear layers $\boldsymbol{\mathrm{W}}_l^1,\boldsymbol{\mathrm{W}}_l^2\in \mathbb{R}^{d\times N_M}$ where $N_M$ is the intermediate size of MLP.  The MLP computation is
\begin{equation}
\setlength{\abovedisplayskip}{8pt}
\setlength{\belowdisplayskip}{8pt}
\begin{aligned}
&\boldsymbol{\mathrm{X}}^{\mathrm{\mathsf{MLP}}}_{l}= \sigma(\boldsymbol{\mathrm{X}}^{\mathsf{MHA}}_{l}\boldsymbol{\mathrm{W}}_l^1)(\boldsymbol{\mathrm{W}}_l^2)^\mathsf{T},
\end{aligned}
\end{equation}
where $\sigma$ is the activation function. 

\begin{figure*}[htb] 
\centering
\includegraphics[width=0.85\textwidth]{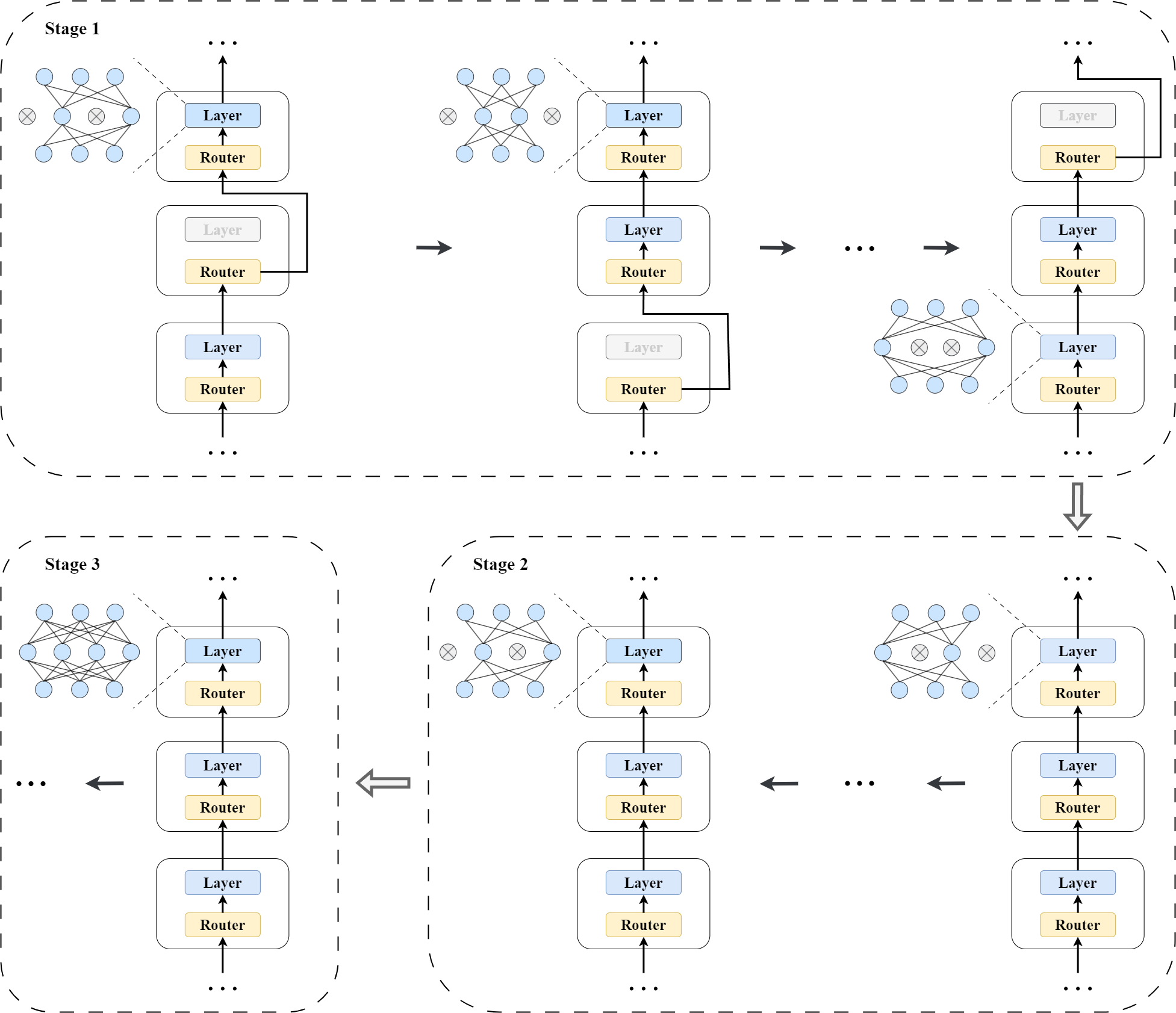}
\caption{Overview of our EST method with practical sampling scheduler. The router takes $\mathbb{I}_L$ as input and determines whether to activate the current layer. In stage 1, we obtain a subnetwork to train by sampling from $N_H, N_M$ and $N_A$ dimensions. In such subnetworks, only some layers are activated and in each activated layer, and only some attention heads and MLP neurons are used. In stage 2, all layers are activated while in each layer still only a subset of the layer is used. In stage 3, the complete model is activated.} 
\label{fig.process} 
\end{figure*}

\subsection{Subnetwork Training via Random Sampling}
\label{subnetwork_training}
Subnetwork training is a training paradigm that trains a subnetwork of the model in each step rather than training the complete model. Let $\Phi$ denote the parameters of the model, $\phi\subset \Phi$ denote the parameters of the subnetwork, and $L$ denote the loss function. Let $f_\phi$ denote the function of the subnetwork, which takes sequence $\boldsymbol{\mathrm{X}}$ as input and outputs the prediction of next tokens. The object of subnetwork  training is formulated as
\begin{equation}
\setlength{\abovedisplayskip}{8pt}
\setlength{\belowdisplayskip}{8pt}
\begin{aligned}
\min_\phi L(f_\phi(\boldsymbol{\mathrm{X}}),\boldsymbol{\mathrm{y}}),
\end{aligned}
\end{equation}
where $\boldsymbol{\mathrm{y}}$ is the ground truth label. 

In our approach, we sample subnetworks randomly from the complete model in each training step. To obtain subnetworks, we sample across three dimensions related to the size of the Transformer model: the number of attention heads $N_H$, the intermediate size of MLP module $N_M$, and the total number of Transformer layers $N_L$.

\textbf{Sampling Attention Heads $N_H$:}
For each MHA module, during the subnetwork training, we randomly sample a subset of heads $\mathbb{I}_H$ for computation at each training step, where $\mathbb{I}_H\subset \{1,2,...,N_H\}$, $|\mathbb{I}_H| = N_Hp_H$ and $p_H$ is the sampling rate. Formally, during the subnetwork training, the output of MHA module is
\begin{equation}
\setlength{\abovedisplayskip}{8pt}
\setlength{\belowdisplayskip}{8pt}
\begin{aligned}
&\boldsymbol{\mathrm{h}}_{l,i} = \mathrm{softmax}\left(\frac{\boldsymbol{\mathrm{X}}_{l-1}\boldsymbol{\mathrm{W}}_{l,i}^Q(\boldsymbol{ 
\mathrm{X}}_{l-1}\boldsymbol{\mathrm{W}}_{l,i}^K)^\mathsf{T}}{\sqrt{d_k}}\right)\boldsymbol{\mathrm{X}}_{l-1}\boldsymbol{\mathrm{W}}_{l,i}^V,\\ 
&\boldsymbol{\mathrm{X}}^{\mathsf{MHA}}_{l}=\frac{1}{p_H}\sum_{i\in \mathbb{I}_H} \boldsymbol{\mathrm{h}}_{l,i}\boldsymbol{\mathrm{W}}_{l,i}^O.
\end{aligned}
\end{equation}
The normalization operation $\frac{1}{p_H}$ is crucial as it ensures that the output distribution is consistent with the complete model. The computational cost of the MHA module in the subnetwork is reduced to a fraction $p_H$ of that in the complete model. The detailed implementation is shown in Appendix~\ref{appendix:implementation_MHA}.

\textbf{Sampling MLP Intermediate Size $N_M$:}
For each MLP module, we sample the intermediate dimension, \emph{i.e.}, the columns of $\boldsymbol{\mathrm{W}}_l^1$ and $\boldsymbol{\mathrm{W}}_l^2$, at each training step. Let $\mathbb{I}_M\subset \{1,2,...,N_M\}$ denote the index of sampled columns where $|\mathbb{I}_M| = N_Mp_M$ and $p_M$ is the sampling rate. We select columns in $\mathbb{I}_M$ from the $\boldsymbol{\mathrm{W}}_l^1$ and $\boldsymbol{\mathrm{W}}_l^2$ to obtain $\hat{\boldsymbol{\mathrm{W}}}_l^1$ and $\hat{\boldsymbol{\mathrm{W}}}_l^2$. The subnetwork's MLP computation is
\begin{equation}
\setlength{\abovedisplayskip}{8pt}
\setlength{\belowdisplayskip}{8pt}
\begin{aligned}
&\boldsymbol{\mathrm{X}}^{\mathsf{MLP}}_{l} = \frac{1}{p_M}\sigma(\boldsymbol{\mathrm{X}}^{\mathsf{MHA}}_{l}\hat{\boldsymbol{\mathrm{W}}}_l^1)(\hat{\boldsymbol{\mathrm{W}}}_l^2)^{\mathsf{T}}.
\end{aligned}
\end{equation}
Similarly, the the output of MLP module requires normalization to ensure the consistency of the output distribution. The computational cost of the MLP module during subnetwork training is reduced to a fraction $p_M$. The detailed implementation is shown in Appendix~\ref{appendix:implementation_MLP}.

\textbf{Sampling Transformer The Number of Layer $N_L$:}
We employ a sampling strategy similar to Stochastic Depth~\cite{stochastic_depth}, randomly skipping some layers of the Transformer model. Let $p_L$ denote the sampling rate. We sample from Transformer layers to obtain a subset $\mathbb{I}_L\subset \{1,2,...,N_L\}$ where  $|\mathbb{I}_L| = N_Lp_L$. For each layer in the Transformer, we compute the layer output as
\begin{equation}
\setlength{\abovedisplayskip}{8pt}
\setlength{\belowdisplayskip}{8pt}
\boldsymbol{\mathrm{X}}_l = \begin{cases}
\boldsymbol{\mathrm{X}}_{l-1} +\mathrm{Layer}_l( \boldsymbol{\mathrm{X}}_{l-1}),  & \text{if $l\in \mathbb{I}_L$} \\
\boldsymbol{\mathrm{X}
}_{l-1}, & \text{if $l\notin \mathbb{I}_L$}
\end{cases}
.
\end{equation}

During subnetwork training, the computational cost of the subnetwork can be reduced to a fraction $p_L$ of the complete model, for only $N_Lp_L$ layers are activated.

\subsection{Evolving  Subnetwork Training}
\label{evolving_training}
In this section, we propose our novel training paradigm for large language models, \textbf{E}volving \textbf{S}ubnetwork \textbf{T}raining~(EST), which applies subnetwork training method in Section~\ref{subnetwork_training}, and progressively increases the size of subnetworks. Finally, train the complete model. 
\begin{definition} Let $T$ denote the total stages of training. A sampling scheduler consists of two parts:~1) $S=(s^{1},...,s^{T})$ that contains the time points of stage transitions, indicating when to increase the size of subnetworks;~2) $P = [(p_H^{1},p_M^{1},p_L^{1}),...,(p_H^{T},p_M^{T},p_L^{T})]$ that contains sampling rates in each stage, indicating how to increase the size of subnetworks.
\end{definition}

EST employs the sampling scheduler to plan the training process. By incrementally increasing the size of subnetworks as the training stages progress and eventually training the complete model, EST achieves training acceleration. The pseudo-code of EST is as Algorithm \ref{alg:EST}.

\begin{algorithm}[!h]
    \caption{Evolving  Subnetwork Training}
    \label{alg:EST}
    \renewcommand{\algorithmicrequire}{\textbf{Input:}}
    \renewcommand{\algorithmicensure}{\textbf{Output:}}
    
    \begin{algorithmic}[1]
        \REQUIRE Dataset $(\mathcal{X},\mathcal{Y})$, sampling scheduler $S=(s^1,...,s^T)$, $P=[(p_H^1,p_M^1,p_L^1),...,(p_H^T,p_M^T,p_L^T)]$.    
        \STATE Randomly initialize the model.
        \FOR[Training stages]{$t=1\to T$} 
            \FOR[Training steps, $s^0=0$]{$k=s^{t-1}\to s^{t}$}
                \STATE Sample $(\mathbf{X},\mathbf{y}) \sim (\mathcal{X},\mathcal{Y})$.
                \STATE Sample $I_L\subset \{1,2,...,N_L\}, |I_L|=p_LN_L$.
                \STATE $\boldsymbol{\mathrm{X}}_0=\mathrm{EMBEDDING}(\boldsymbol{\mathrm{X}})$.
                \FOR[Transformer layers]{$l=1\to N_L$}
                \IF{$l\notin \mathbb{I}_L$}
                    \STATE $\boldsymbol{\mathrm{X}}_l = \boldsymbol{\mathrm{X}}_{l-1}$.
                    \STATE Continue.
                \ENDIF
                \STATE Sample $I_H\subset \{1,2,...,N_H\}, |I_H|=p_HN_H$.
                \STATE Sample $I_M\subset \{1,2,...,N_M\}, |I_M|=p_MN_M$.
                \STATE Compute $\boldsymbol{\mathrm{X}}^{\mathsf{MHA}}_{l}$ condition on $\mathbb{I}_M$.
                \STATE Compute $\boldsymbol{\mathrm{X}}^{\mathsf{MLP}}_{l}$ condition on $\mathbb{I}_H$.
                \STATE $\boldsymbol{\mathrm{X}}_{l}=\boldsymbol{\mathrm{X}}_{l-1}+\boldsymbol{\mathrm{X}}^{\mathrm{MLP}}_{l}$.
                \ENDFOR
                \STATE $\boldsymbol{\mathrm{\hat{y}}}=\mathrm{PROJECTION}(\boldsymbol{\mathrm{X}}_{N_L})$.
                \STATE Compute loss with $L(\boldsymbol{\mathrm{\hat{y}}},\boldsymbol{\mathrm{y}})$ .
                \STATE Backward and optimize.
            \ENDFOR
        \ENDFOR
    \end{algorithmic}
\end{algorithm}

\textbf{Practical Sampling Scheduler:}
In this paper, we won't dive into complex sampling schedulers. For convenience, we use a kind of three-stage sampling scheduler in practice. Specifically, our sampling scheduler consists of the following three stages:
\begin{itemize}
  \item \textbf{Stage 1}: In this stage, we sample from all three dimensions to achieve the highest acceleration ratio. That is, $0<p_H<1$, $0<p_M<1$, $0<p_L<1$.
  \item \textbf{Stage 2}: In this stage, we stop sampling from the Transformer layers, ensuring that the number of activated layers in the subnetwork is consistent with the complete model, while continuing sampling from the MHA and MLP modules. That is, $0<p_H<1$, $0<p_M<1$, $p_L=1$. Additionally, in this stage, we keep $p_H$ and $p_M$ consistent with the first stage.
  \item \textbf{Stage 3}: In the final stage, train the complete model, where $p_H=1$, $p_M=1$, $p_L=1$.
\end{itemize}

The process of EST with this kind of sampling scheduler is illustrated in Figure~\ref{fig.process}.

\textbf{Training Cost Saving:}
Assuming that, under the condition of equal training steps, the model trained by EST achieves the same performance as the original model, EST can indeed achieve a reduction in training cost. This is because the cost of training subnetworks is smaller than that of training the complete model. 

For ease of analysis, we calculate how much training cost is saved by EST with the practical sampling scheduler. Let $C_H, C_M$ denote the cost of each MHA module and MLP module, neglecting other modules like Layer Normalization~\cite{layernorm} as their cost is relatively small compared to MHA and MLP. So the training cost of a single training step in each stage is formulated as
\begin{equation}
\setlength{\abovedisplayskip}{8pt}
\setlength{\belowdisplayskip}{8pt}
\begin{aligned}
&C_1 = N_Lp_L(p_H C_H+p_M C_M),\\
&C_2 = N_L(p_H C_H+p_M C_M),\\
&C_3 = N_L(C_H+C_M).
\label{stage_cost}
\end{aligned}
\end{equation}
Based on the training steps for each stage, the total training cost can be calculated. Let $r_1,r_2,r_3$ denote training steps of each stage, respectively. The total training cost is formulated as
\begin{equation}
\begin{aligned}
C_{EST} &= r_1C_1+r_2C_2+r_3C_3\\
&=(r_1N_Lp_Lp_H+r_2N_Lp_H+r_3N_L)C_H\\
&+(r_1N_Lp_Lp_M+r_2N_Lp_M+r_3N_L)C_M.
\end{aligned}
\end{equation}
On the other hand,  the training cost of training the complete model is
\begin{equation}
\begin{aligned}
C_{original} &= (r_1+r_2+r_3)N_L(C_H+C_M).
\end{aligned}
\end{equation}

For a more intuitive illustration, Table~\ref{tab:cost_saving} shows the  training cost under specific configurations that $p_H=p_M=p_L=0.5$, $r_1=r_2=r_3=r$ compared with cost of training the original model through naive training method.  Under such configurations, EST can save $41.7\%$ of training cost.  
\begin{table}[htbp]
  \centering
  \caption{An intuitive example of training cost saving. The real world wall time saving is shown in Appendix~\ref{appendix:wall_time}}.
  \label{tab:cost_saving}
  \begin{tabular}{lcc}
    \toprule
    Stages&EST & Original \\
    \midrule
    Stage 1 & $0.25rN_L (C_H+C_M)$ & $rN_L (C_H+C_M)$\\
    Stage 2 & $0.5rN_L (C_H+C_M)$ & $rN_L (C_H+C_M)$\\
    Stage 3 & $rN_L (C_H+C_M)$ & $rN_L (C_H+C_M)$\\
    \midrule
    Total & $1.75rN_L (C_H+C_M)$ & $3rN_L (C_H+C_M)$\\
    Saving & $1.25rN_L (C_H+C_M)$ & 0\\
    \bottomrule
  \end{tabular}
\end{table}

%% file: sections/experiment.tex
\section{Experiments}
\label{Experiment}

In this section, we first present our main results with GPT2 model on the in-domain pre-train task and out-domain downstream tasks in Section~\ref{gpt2_experiment}. In addition, to show the scalability of our approach, we conduct experiments with TinyLlama model in Section~\ref{llama2_experiment}.

\subsection{Main Results with GPT2}
\label{gpt2_experiment}
\textbf{Experiment Setup:} We conduct experiments with GPT2-base model, which has 117M parameters in total, pre-trained on OpenWebText dataset~\cite{gpt2} with AdamW optimizer~\cite{adamw} from scratch. The batch size is set to 512 and the sequence length is 1024. The total training step is 150k. For the downstream performance, we experiment on three tasks: GLUE~\cite{glue}, SQuAD~\cite{squad} , and LAMBADA~\cite{lambada}. 

For GPT2-base model, the practical sampling scheduler is set to $S=(20k,70k,150k)$ and $P=[(0.5,0.5,0.5),(0.5,0.5,1),(1,1,1)]$, which saves $26.7\%$ computation cost of training.

We choose Staged Training~\cite{shen2022staged}, which is a kind of incremental training method and has two stages, as a baseline. In stage 1, Staged Training method trains the model with half hidden size. At the end of stage 1, expand the parameters of the model. In stage 2, train the complete model. The stage transition occurs at step 50k, and this baseline saves $16.7\%$ of the training FLOPs.

For another baseline MSG~\cite{MSG}, due to the inability of this method to simultaneously expand attention heads and intermediate sizes, the training stage settings in the MSG baseline differ slightly from our EST method:
\begin{itemize}
  \item \textbf{Stage1 (0-20k)}: Utilizing a model with half the number of layers, attention heads, and intermediate size. 
  \item \textbf{Stage2 (20k-40k)}: Restoring the number of layers and using a model with half the attention heads and intermediate size. 
  \item  \textbf{Stage3 (40k-70k)}: Restoring the attention heads and using a model with half the intermediate size. 
  \item  \textbf{Stage4 (70k-150k)}: Training the complete model.
\end{itemize}

\textbf{Main Results:}  
We compare EST with the naive training method that trains the complete model, Staged Training method and MSG method. The results are as the Table~\ref{gpt2_main_results}. Our approach EST saves $26.7\%$ of training FLOPs and leads to 1.22x speed up of the wall clock training time without increasing the loss on the validation dataset. The loss curve of GPT2 trained by EST can be found in Appendix~\ref{appendix:gpt_detail}. Additionally, we find that the model trained by EST has better downstream performance, indicating that EST also enhances the generalization of GPT2 model.  However, despite saving 16.7\% of the training FLOPs, the performance of the model obtained by Staged Training cannot match the original model. Compared to MSG method, our EST approach achieves higher acceleration effects and delivering superior model performance.
 
\begin{table*}[!t]
\vspace{-0.3cm}
\caption{Main results of the experiment with GPT2 model. We choose Staged Training~\cite{shen2022staged} and MSG~\cite{MSG} baselines. Loss is evaluated on the validation dataset. For metrics, we use accuracy and F1 score for SQuAD, and accuracy for LAMBADA. The detailed results of GLUE are in Appendix~\ref{appendix:gpt_detail}.}
\label{gpt2_main_results}
\vskip 0.15in
\begin{center}
\begin{sc}
\begin{tabular*}{0.75\linewidth}{lccc}
\toprule 
  &Wall Clock Time(hours)&Speed Up& Saving FLOPs \\
\midrule
Original&185.0&1x &0\\
Staged Training&173.1&1.06x&16.7\%\\
MSG &160.8&1.16x & 24.4\% \\
\midrule
EST &\textbf{151.6}&\textbf{1.22x}& \textbf{26.7\%}\\
\bottomrule
\end{tabular*}
\begin{tabular*}{0.75\linewidth}{lcccc}
\toprule 
  &Average GLUE & SQuAD & LAMBADA& Loss \\
\midrule
Original&79.84 & 66.74/77.06&29.44&3.06\\
Staged Training&73.82 &61.65/72.71&28.99&3.15\\
MSG &79.88&62.05/71.89&29.06&3.13 \\
\midrule
EST&\textbf{80.66}&\textbf{67.14}/\textbf{77.15}&\textbf{32.01}&\textbf{3.05}\\
\bottomrule
\end{tabular*}
\end{sc}
\end{center}
\vskip -0.1in
\end{table*}

\textbf{Effect of Sampling Scheduler:}
We also conduct experiments to evaluate the effect of different sampling schedulers. Besides our practical sampling scheduler, we evaluate model performance on five different sampling schedulers:
\begin{itemize}
    \item EST-ONE-STAGE: $S=(150k)$, $P=[(0.5,0.5,1)]$.
    \item EST-TWO-STAGE-A: $S=(50k,150k)$, $P=[(0.5,0.5,1),(1,1,1)]$.
    \item EST-TWO-STAGE-B: $S=(70k,150k)$, $P=[(0.5,0.5,1),(1,1,1)]$.
    \item EST-TWO-STAGE-C: $S=(90k,150k)$, $P=[(0.5,0.5,1),(1,1,1)]$.
    \item EST-THREE-STAGE: $S=(20k,70k,150k)$, $P=[(0.5,0.5,0.5),(1,1,0.5),(1,1,1)]$.
\end{itemize}
    Among these five sampling schedulers, EST-ONE-STAGE is exactly the stage 2 of the practical sampling scheduler and does not train the complete model at all.  EST-TWO-STAGE-A, EST-TWO-STAGE-B, and EST-TWO-STAGE-C skip the stage 1 of the practical sampling scheduler and have different stage transition points.  EST-THREE-STAGE, on the other hand, modifies the stage 2 of the practical sampling scheduler by disabling the sampling from MHA and MLP and enabling the sampling from the number of layers. The results are shown in Table~\ref{gpt2_scheduling_results}. 
    
We find that our three-stage practical sampling scheduler saves more training cost than two-stage sampling schedulers with comparable model performance. On the other hand, results of EST-ONE-STAGE and EST-TWO-STAGE-C indicate that a long enough stage 3 is vital. In addition, the experiment on EST-THREE-STAGE sampling scheduler indicates that, in stage 2, sampling from MHA modules and MLP modules performs better than sampling from  layers.

\begin{table*}[!t]
\vspace{-0.2cm}
\caption{Ablation study on different sampling schedulers. We find that our proposed practical sampling scheduler strikes a good balance between model performance and FLOPs saving. Compared to other types of schedulers, our practical sampling scheduler performs the best with the same training cost.}
\label{gpt2_scheduling_results}
\vskip 0.15in
\begin{center}
\begin{sc}
\begin{tabular*}{0.9\linewidth}{lccccc}
\toprule
  &Saving FLOPs &Loss & Average GLUE & SQuAD & LAMBADA \\
\midrule
Original& 0&3.06& 79.84 & 66.74/77.06&29.44\\
\midrule
EST-one-stage & \textbf{50.0\%} &3.36&77.78&63.95/74.65&26.57 \\
EST-two-stage-A &16.7\%&\textbf{3.04} & \textbf{81.05}&\textbf{67.39/77.76} &29.59\\
EST-two-stage-B & 23.3\% &3.06 & 80.17&67.18/77.51 &29.71\\
EST-two-stage-C &30.0\%&3.09&80.41& 66.55/77.01& 28.68\\
EST-three-stage&26.7\%&3.07&79.47& 65.73/76.22& 29.67\\
\midrule
EST & 26.7\% &3.05 &80.66 &67.14/77.15&\textbf{32.01}\\
\bottomrule
\end{tabular*}
\end{sc}
\end{center}
\vskip -0.1in
\end{table*}

\subsection{Scale-up to TinyLlama}
\label{llama2_experiment}
\textbf{Experiment Setup:} 
We pre-train a 1.1B TinyLlama model with 22 layers on the subset of SlimPajama dataset~\cite{cerebras2023slimpajama} and Starcoder dataset~\cite{li2023starcoder} from scratch, using AdamW optimizer. The batch size is set to 1024 and the sequence length is 2048. The total training step is 60k, containing 130B tokens in total. We report the validation loss on SlimPajama dataset and downstream performance on GPT4All~\cite{gpt4all} benchmarks. GPT4All contains seven different datasets, evaluating the few-shot common sense reasoning ability of models.

For TinyLlama model, the practical sampling scheduler is set to $S=(10k,25k,60k)$ and $P=[(0.5,0.5,0.5),(0.5,0.5,1),(1,1,1)]$, which saves $25.0\%$ computation cost of training.

\textbf{Main Results:}
We demonstrate the scalability of EST on TinyLlama in Table~\ref{llama2_main_results}.  Compared with the original model trained by the naive training method, EST method saves $25.0\%$ of training FLOPs and leads to 1.22x speed up of wall clock training time, with comparable loss. The loss curve of TinyLlama trained by EST can be found in Appendix~\ref{appendix:llama_detail}. In addition, model generalization performance, measured by the average score of GPT4All, is improved by EST. 

\begin{table*}[htb]
\vspace{-0.5cm}
\caption{Main results of experiment with TinyLlama model. Loss is evaluated on the validation dataset. The detailed results of GPT4All are shown in 
Appendix~\ref{appendix:llama_detail}.}
\label{llama2_main_results}
\begin{center}
\begin{sc}
\begin{tabular}{lccccc}
\toprule
 &Wall Clock Time(hours)&Speed Up& Saving FLOPs&Loss&GPT4All\\
\midrule
Original&192.8&1x&0&\textbf{2.64}&42.40\\
\midrule
EST&\textbf{158.2}&\textbf{1.22x}&\textbf{25\%}&2.65&\textbf{42.79}\\
\bottomrule
\end{tabular}
\end{sc}
\end{center}
\end{table*}

%% file: sections/theory.tex
\section{Theoretical Studies}
In this section, we aim to answer two key questions: 1) Why can EST method save training cost without compromising model performance? 2) Why do models trained using the EST method exhibit better generalization performance?
We study the training dynamics of EST in Section~\ref{est_dynamics} to answer question 1 and study the loss landscape of models trained using the EST method in Section~\ref{est_loss_landscape} to answer question 2.
\label{Theoretical Studies}
\subsection{Why Can EST Save Training Cost without Compromising Model Performance}
\label{est_dynamics}
In general, EST enhances the dynamics of model training, resulting in a steeper loss curve and faster loss descent compared to the naive training approach, so it can save training cost without compromising model performance. To provide a more specific explanation, we first need to introduce two important properties in previous incremental training methods~\cite{shen2022staged}: \emph{loss-preserving property} and \emph{training-dynamics-preserving property}. Subsequently, we point out that it is precisely because EST breaks away from these two properties that it can achieve savings in model training cost in a broad range of scenarios.

\textbf{Loss-preserving Property:} The loss-preserving property implies that during a transition in the training stages, the models before and after the transition should represent the same function, resulting in identical loss.

\textbf{Training-dynamics-preserving Property:} Intuitively, the training-dynamics-preserving property means that in the final stage of incremental training, the loss curve should match that of the target model.

\textbf{Why Should Break Away from These Properties:}
In incremental training methods, maintaining these two properties is to ensure the feasibility of extending the parameters of a smaller model as the initialization parameters for the target model. However, to maintain these two properties while achieving the goal of saving training cost, incremental training methods require expanding the parameters to the size of the target model early in the training process~\cite{shen2022staged}. Therefore, when the model training requires a sufficient number of training steps, incremental training methods can actually save very little in training cost. EST breaks away from these two properties, alleviating this issue.

\textbf{Break Away from Loss-preserving Property:}
EST method dose not maintain the loss-preserving property. During stage transitions when increasing the size of the subnetworks, the loss experiences a sudden drop compared to the previous stage.

Due to the equivalence between the random sampling of subnetworks and  Structural Dropout~\cite{structural_dropout}, we can theoretically demonstrate using Dropout theory. Intuitively, training subnetworks via random sampling implicitly introduces a regularization term to the loss function, and the increase in subnetwork size reduces this regularization term, resulting in a sudden drop in loss.

\textbf{Break Away From Training-dynamics-preserving Property:}
EST method does not maintain the training-dynamics-preserving property. In the final stage of training, the model trained with EST exhibits better training dynamics compared to the target model obtained through naive training method. 

Intuitively, EST method is equivalent to the use of Structural Dropout in earlier stages. Early Dropout pushes model parameters into a flatter region of the loss landscape. Even in the final stage when training the complete model, the model parameters can still maintain flatness, reducing the difficulty of parameter optimization and accelerating the descent of the loss. 

\begin{figure}[!t]
  \centering
  \vspace{-0.3cm}
 \includegraphics[width=\linewidth,trim=0cm 1cm 0cm 1cm, clip]{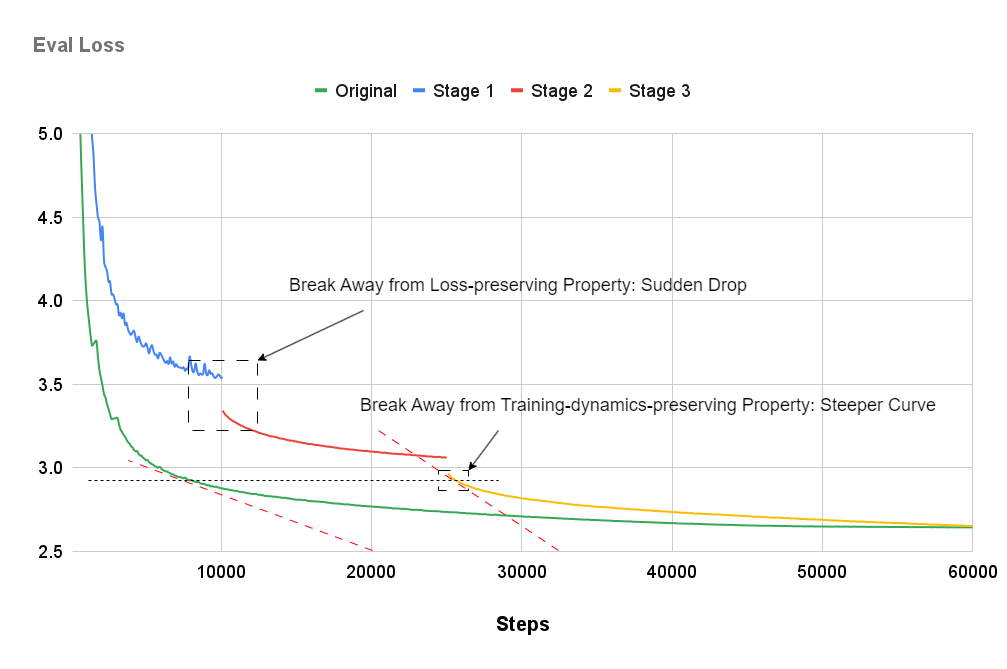}
 \vspace{-0.5cm}
  \caption{Loss curves of EST compared with the original training method. }
  \label{fig:properties}
   \vspace{-0.30cm}
\end{figure}

\textbf{Towards Training Cost Saving:} How does the disruption of these two properties affect the efficiency of model training? We compare the loss curves of TinyLlama trained by EST and the original training method in Figure~\ref{fig:properties}.

Firstly, we observe that during stage transitions, the loss drops sharply, which is the result of breaking the loss-preserving property, providing a better starting point for each stage. On the other hand, we notice that in stage 3, the slope of the EST loss curve is greater than the slope of the original loss curve under the same loss condition. This is a direct result of breaking the training-dynamics-preserving property, which is the key to the success of EST. If the training-dynamics-preserving property holds, the loss curve in stage 3 will be parallel to the loss curve of the original training method, and thus, the two curves will not intersect, eliminating the possibility of saving training costs. It is precisely because the EST method improves the dynamics of model training that it leads to savings in training costs without sacrificing model performance.
\subsection{Why Can EST Benefit Model Generalization} 
\label{est_loss_landscape}
In Section~\ref{Experiment}, we observe that the EST method not only achieves comparable loss on the pre-training dataset compared to the naive training method, but also brings some improvement in downstream tasks. This indicates an enhancement in the model's generalization performance.

\citet{liu2023same} investigate the phenomenon where different models exhibit significant differences in downstream tasks under the same loss on the pre-training dataset. Furthermore, they find a strong correlation between the model's generalization ability and the trace of the Hessian matrix of the loss function with respect to the model parameters.

The process of sampling subnetworks in the EST method is equivalent to Structural Dropout, which can effectively reduce the trace of the Hessian matrix in the early stages. However, more importantly, we find that even in the final stage of training the complete model, the Hessian matrix still maintains a relatively small trace until the end of training. This contributes to the better generalization performance of the final model obtained through EST. 

In Table \ref{tab:generalization_hessian}, we demonstrate that GPT2 models trained through EST exhibit a smaller trace of the Hessian matrix and stronger generalization performance compared to models obtained through naive training method. Here $\mathrm{Tr}[\nabla^2L(\phi)]$ denotes the trace of Hessian matrix of the loss function with respect to the model parameters.

\begin{table}[!t]
  \centering
\vspace{-0.4cm}
  \caption{The trace of Hessian matrix and the average GLUE score of model trained through naive training method and through EST method. With almost the same pre-training loss, EST method has higher GLUE score for smaller trace of Hessian matrix.}
\vspace{0.2cm} 
  \label{tab:generalization_hessian}
  \begin{tabular}{lccc}
    \toprule
    &Loss & GLUE & $\mathrm{Tr}[\nabla^2L(\phi)]$ \\
    \midrule
    Original&3.06&73.89& 11763\\
    \midrule
    EST&3.05&74.66& 2510\\
    \bottomrule
  \end{tabular}
  \vspace{-0.5cm}
\end{table}

%% file: sections/conclusion.tex
\section{Conclusion}
\label{Conclusion}

Our goal is to achieve more efficient training for large language models. We propose a novel training method, Evolving Subnetwork Training~(EST), which operates subnetwork training via random sampling and uses sampling scheduler to plan the process of training incrementally. Our approach enhances the efficiency of model training on GPT2 and TinyLlama models, saving 26.7\% and 25.0\% of training FLOPs respectively, with comparable pre-training performance. In addition, EST benefits the generalization ability of both GPT2 and TinyLlama, evaluated by several downstream tasks. We also provide intuitive theory studies, demonstrating the feasibility and superiority of EST. Through theoretical analysis, we find that the efficient training dynamics of EST comes from the flatness of parameters. This insight may inspire other efficient training methods. In future works, we aim to provide the theoretical support for the design of sampling schedulers, to apply EST for training on even larger models. Additionally, since EST essentially samples for matrix multiplication, it can be applied not only to Transformer models but also to models like Mamba~\cite{gu2023mamba}. We will conduct experiments on other types of models to broaden the application scope of EST.

%% file: sections/appendix.tex
\newpage
\appendix
\onecolumn
\section{Additional Experiment Details}
\label{appendix:add_detail}
\subsection{Details for GPT2 Experiment}
\label{appendix:gpt_detail}
\begin{figure*}[h]
  \centering

  \subfigure{
    \includegraphics[width=0.48\linewidth]{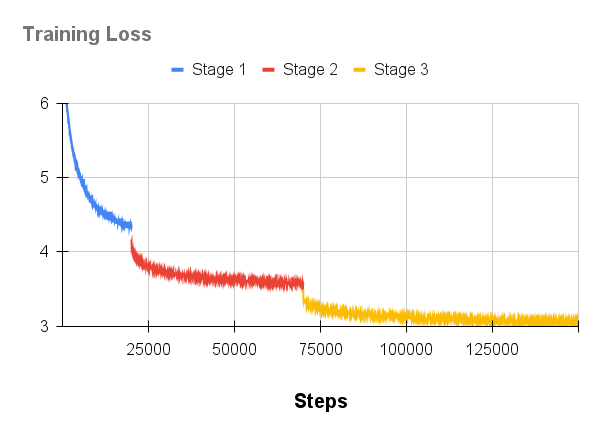}
   }
    \hfill
  \subfigure{
    \includegraphics[width=0.48\linewidth]{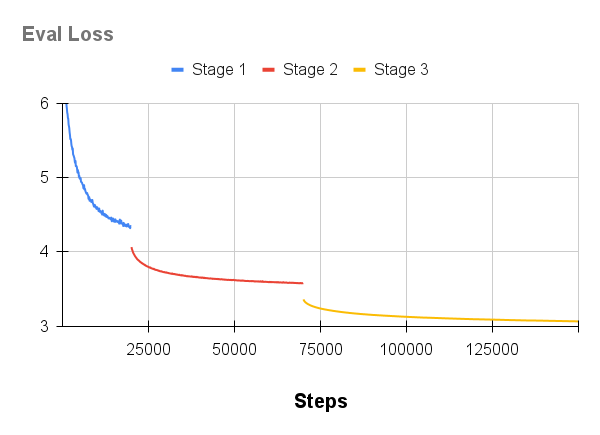}
  }

  \caption{Training and evaluation loss of EST training with GPT2-base model.}
  \label{fig:gpt_loss}
\end{figure*}

\textbf{Details for Pre-training:} We pre-train 117M GPT2 model on OpenWebText dataset from scratch, using AdamW optimizer. Batch size is set to 512 and each example contains 1024 tokens. The initial learning rate is set to $6\times10^{-4}$, followed by a linear learning rate decay. The pre-training loss curves of EST method on training dataset and validation dataset are as Figure~\ref{fig:gpt_loss}.

\textbf{Details for GLUE benchmark}:
The detailed scores evaluated on GLUE are as Table~\ref{tab:gpt2_glue}. CoLA is measured by Matthews correlation and accuracy. STS-B is measured by Pearson/Spearman correlation. MRCP and QQP are measured by accuracy and F1 score. Others are measured by accuracy.
\begin{table*}[htbp]
\caption{Detailed GLUE scores of GPT2-base model.}
\label{tab:gpt2_glue}
\vskip 0.15in
\begin{center}
\begin{sc}
\begin{tabular*}{0.85\linewidth}{lcccccccc}
\toprule
  &Saving FLOPs&CoLA &SST-2 & MRPC &STS-B \\
\midrule
Original&0& 43.21/77.27&90.02& 80.15/86.39 & 86.63/86.25\\
Staged Training&16.7\%&14.50/65.68&87.15&76.25/84.43&83.97/83.76 \\
\midrule
EST-one-stage&50.0\%&37.27/74.88&89.91&76.23/83.81&82.62/82.36\\
EST-two-stage-A&16.7\% &44.61/78.04&91.51 & 83.58/88.39&87.11/86.97\\
EST-two-stage-B&23.3\% & 43.26/77.09 &91.06 & 80.39/86.44&86.56/86.57 \\
EST-two-stage-C&30.0\% &45.70/78.14&89.79&80.15/86.43& 86.71/86.52\\
EST-three-stage&26.7\%&42.66/76.22&91.28&80.16/86.48&84.33/84.12\\
\midrule
EST &26.7\% &  45.88/78.04 &91.51 &80.88/87.13&86.69/86.55\\
\bottomrule
\end{tabular*}
\begin{tabular*}{0.85\linewidth}{lcccccc}
\toprule
 &Saving FLOPs& QQP &MNLI(m/mm)&QNLI&RTE\\
\midrule
Original&0&89.50/85.83&79.14/79.59&86.07&67.87\\
Staged Training&16.7\%&86.91/82.36&75.45/75.79&82.44&61.01 \\
\midrule
EST-one-stage&50.0\%&89.05/85.26&77.81/78.29&86.16&67.51\\
EST-two-stage-A&16.7\%&89.39/85.70&80.24/80.47&86.94&70.76\\
EST-two-stage-B&23.3\% &89.55/85.80&80.02/80.73&86.49&68.23\\
EST-two-stage-C&30.0\% & 89.40/85.82&80.36/80.08&86.88&69.31\\
EST-three-stage&26.7\%&89.25/85.49&79.01/80.13&86.25&67.79\\
\midrule
EST &26.7\% &89.27/85.53&80.38/80.81&86.99&68.95\\
\bottomrule
\end{tabular*}
\end{sc}
\end{center}
\vskip -0.1in
\end{table*}

\subsection{Details for TinyLlama  Experiment}
\label{appendix:llama_detail}
\begin{figure*}[h]
  \centering

  \subfigure{
    \includegraphics[width=0.48\linewidth]{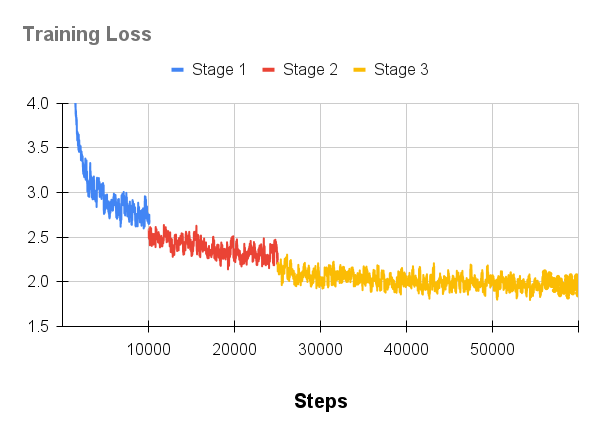}
   }
    \hfill
  \subfigure{
    \includegraphics[width=0.48\linewidth]{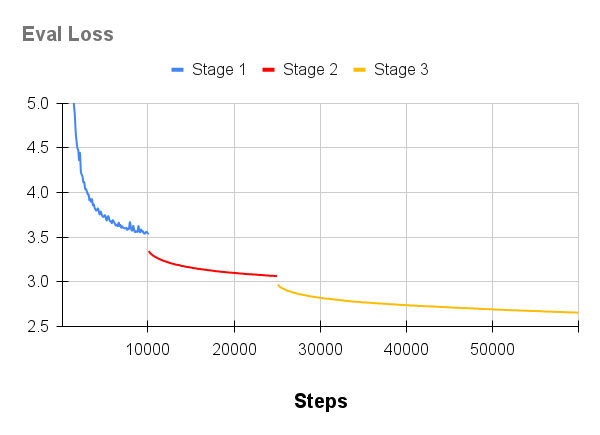}
  }

  \caption{Training and evaluation loss of EST training with GPT2-base model.}
  \label{fig:llama_loss}
\end{figure*}

\textbf{Details for Pre-training:}
We pre-train TinyLlama 1.1B model on the subset of SlimPajama dataset and Starcoder dataset, consisting of 130B tokens. We use AdamW optimizer, and the max learning rate is set to $4\times10^{-4}$ with 2000 warm-up steps, followed by a cosine learning rate decay. The pre-training loss curves of EST method on training dataset and validation dataset are as Figure~\ref{fig:llama_loss}.

\textbf{Details for GPT4All benchmark:}
The detailed scores evaluated on GPT4All are as Table~\ref{tab:llama_gpt4all}. 
\begin{table*}[htbp]
\caption{Detailed GPT4All scores of TinyLlama model.}
\label{tab:llama_gpt4all}
\vskip 0.15in
\begin{center}
\begin{sc}
\begin{tabular*}{\linewidth}{lcccccccc}
\toprule
  &Saving FLOPs&HellaSwag&Obqa&WinoGrande&$\mathrm{ARC_c}$&$\mathrm{ARC_e}$&Boolq&Piqa\\
\midrule
Original&0& 33.54&29.40&50.51&23.04&38.55&59.60&62.13\\
\midrule
$\mathrm{EST}$ &25.0\% & 33.40&27.20&52.88&23.29&38.93&61.16&62.68\\
\bottomrule
\end{tabular*}
\end{sc}
\end{center}
\end{table*}

\subsection{Details for Wall Time Saving}
\label{appendix:wall_time}
We test the efficiency of the EST method on GPT2 and TinyLlama models and assess the real acceleration effects. We will analyze the wall time overhead of each module and the time overhead under different training setups. Here we mainly analyze the impact of sampling on the MHA and MLP modules. These two modules involve matrix multiplication, and our sampling alters the size of these matrices. Since matrix multiplication is parallelized on GPUs, it's challenging to intuitively calculate the actual acceleration effect. For both GPT2-base and  TinyLlama 1.1B model, we investigate the impact of different batch sizes on training speed when using Distributed Data Parallel (DDP).  For simplicity, we discuss the practical sampling scheduler in Table~\ref{tab:cost_saving}. We use A100 80GB GPU to test both GPT2 model and TinyLlama model.

For the GPT2 model, the actual acceleration effects are as Table~\ref{tab:gpt_time}. For the TinyLlama 1.1B model, the actual acceleration effects are as Table \ref{tab:llama1.1B_time}. In these two tables, GPU time refers to the time spent on forward computations for each module or layer on the GPU. Total time indicates the overall time cost for each training step, including both forward and backward computation.

\begin{table*}[htbp]
\caption{Wall time overhead of GPT2 model. Each example contains 1024 tokens. }
\label{tab:gpt_time}
\vskip 0.15in
\begin{center}
\begin{sc}
\begin{tabular*}{0.92\linewidth}{lcccc}
\toprule
  Micro Batch Size	&8&16&32&48 \\
\midrule 
GPU time~(ms) of  MHA&2.51&4.84&9.48&14.12\\
GPU time~(ms) of  EST MHA&1.41&2.57&4.92&7.27\\
\midrule
GPU time~(ms) of  MLP&0.92&1.78&3.38&5.01\\
GPU time~(ms) of  EST MLP&0.59&1.05&1.95&2.86\\
\midrule
GPU time~(ms) of Transformer Layer&3.56&6.86&13.31&19.77\\
GPU time~(ms) of  EST Transformer Layer&2.12&3.86&7.34&10.79\\
\midrule
Total time~(ms) of Stage 1 training step &182.49&278.62&447.91&525.55\\
Total time~(ms) of Stage 2 training step &210.59&309.27&497.37&721.37\\
Total time~(ms) of Stage 3 (original) training step &211.05&388.46&646.85&1065.23\\
\bottomrule
\end{tabular*}
\end{sc}
\end{center}
\vskip -0.1in
\end{table*}

\begin{table*}[htbp]
\caption{Wall time overhead of TinyLlama 1.1B model. Each example contains 2048 tokens. }
\label{tab:llama1.1B_time}
\vskip 0.15in
\begin{center}
\begin{sc}
\begin{tabular*}{0.92\linewidth}{lcccc}
\toprule
  Micro Batch Size& 1 &2 & 4 &8 \\
\midrule 
GPU time~(ms) of  MHA& 1.43&2.10&3.93&8.14\\
GPU time~(ms) of  EST MHA&1.45&1.63&2.24&4.11\\
\midrule
GPU time~(ms) of  MLP& 0.39&0.69&1.43&2.87\\
GPU time~(ms) of  EST MLP&0.44&0.61&0.98&1.60\\
\midrule
GPU time~(ms) of Transformer Layer& 2.05&3.12&5.99&11.88\\
GPU time~(ms) of  EST Transformer Layer&2.50&2.57&3.85&6.84\\
\midrule
Total time~(ms) of Stage 1 training step &288.37&228.66&360.24&501.61\\
Total time~(ms) of Stage 2 training step &374.05&296.95&399.80&659.46\\
Total time~(ms) of Stage 3 (original) training step &274.23&336.71&555.82&915.40\\
\bottomrule
\end{tabular*}
\end{sc}
\end{center}
\vskip -0.1in
\end{table*}

The final speedup rate is not $4\times$ for stage1 and not $2\times$ for stage 2 due to two reasons: (1) GPUs compute matrix multiplication in parallel, so the time consumption is not directly proportional to the number of rows or columns of the matrix; (2) In addition to GPU computation time, there is also high memory access overhead during model training. 
As the batch size increases, the bottleneck of training gradually shifts from memory access to computation, resulting in an increase in the speedup, and the speedup on GPU time gradually approaches $2\times$.

\section{Implementation Details}
\label{appendix:implementation}
\begin{figure*}[htb] 
\centering
\includegraphics[width=0.65\textwidth]{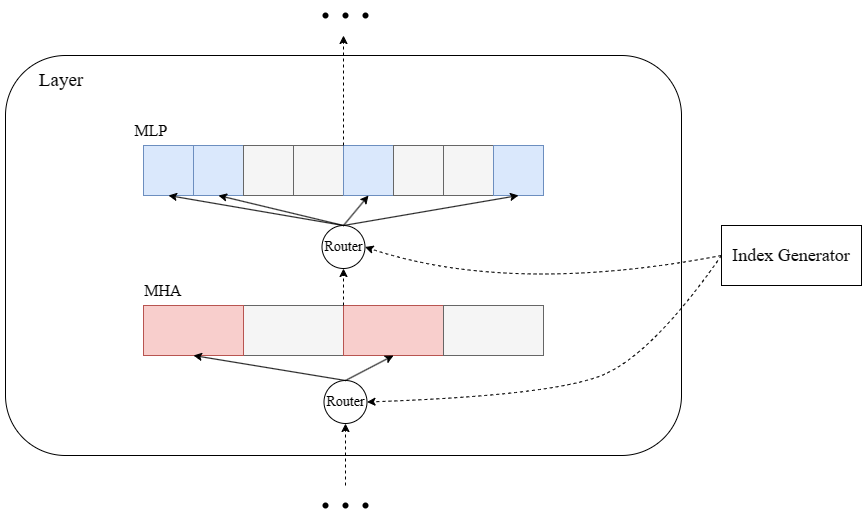}
\caption{Computation in each Transformer layer during subnetwork training.} 
\label{fig.router} 
\end{figure*}

Among the three sampling methods we use, sampling for the number of Transformer layers is straightforward and will not be elaborated. However, the sampling operation for the dimensions of MHA and MLP modules within each layer is more complex. This will be detailed here. The operation within each Transformer layer can be illustrated as Figure~\ref{fig.router}. The index generator generates indexes $\mathbb{I}_H$, $\mathbb{I}_M$ and $\mathbb{I}_L$. The router before each module takes $\mathbb{I}_H$ or $\mathbb{I}_M$ as input and activates the corresponding part of the module.

\subsection{Implementation of Sampling for MHA module}
\label{appendix:implementation_MHA}
\begin{figure*}[htb] 
\centering
\includegraphics[width=0.9\textwidth]{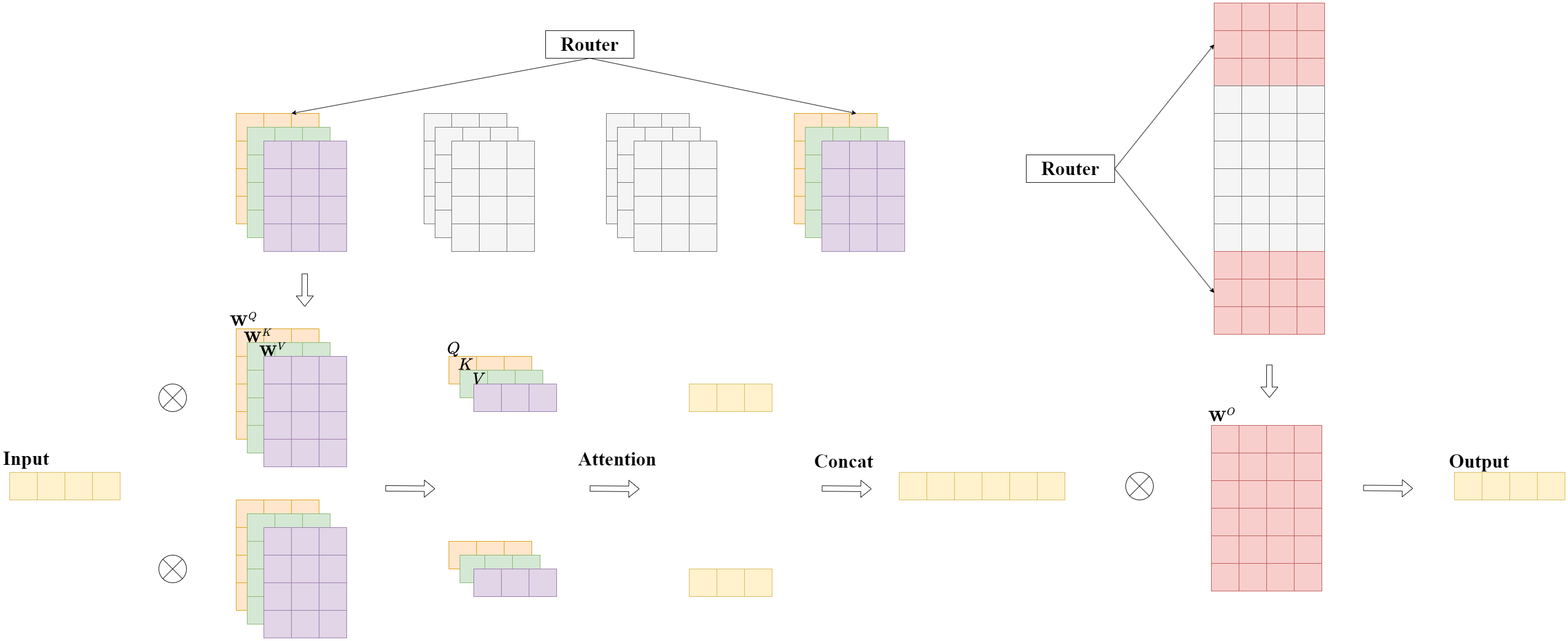}
\caption{The detailed implementation of sampling for MHA module.} 
\label{fig.mha} 
\end{figure*}

For the MHA module, we sample a subset of heads for computation. Specifically, this involves sampling along the dimensions of the output projection matrices $\boldsymbol{\mathrm{W}}^Q$,$\boldsymbol{\mathrm{W}}^K$,$\boldsymbol{\mathrm{W}}^V$, and selecting the corresponding input dimensions in the output matrix $\boldsymbol{\mathrm{W}}^O$. The detailed process is illustrated in Figure~\ref{fig.mha}.
\subsection{Implementation of Sampling for MLP module}
\label{appendix:implementation_MLP}
\begin{figure*}[htb] 
\centering
\includegraphics[width=0.7\textwidth]{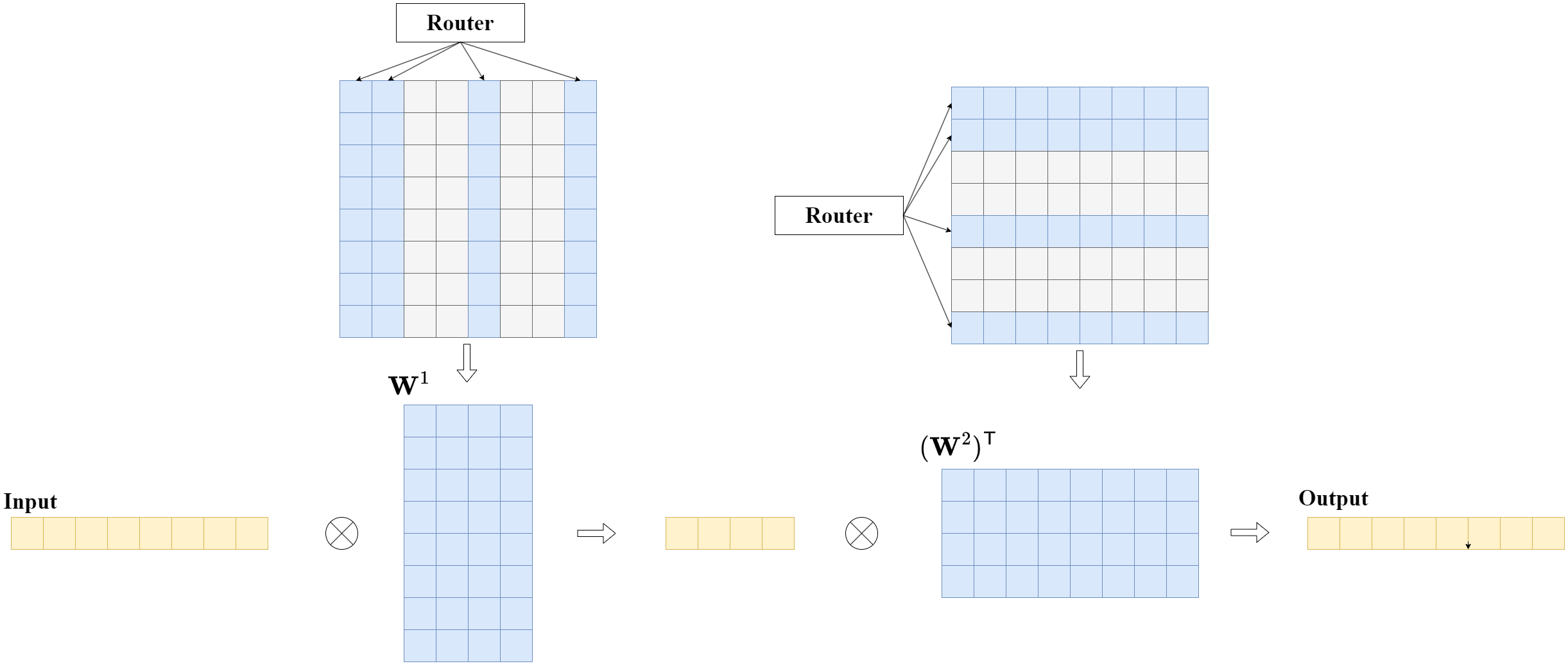}
\caption{The detailed implementation of sampling for MLP module.} 
\label{fig.mlp} 
\end{figure*}

For the MLP module, we sample columns from $\boldsymbol{\mathrm{W}}^1$ and $\boldsymbol{\mathrm{W}}^2$ for computation. The detailed process is illustrated in Figure~\ref{fig.mlp}.

Unlike in Deja Vu~\cite{liu2023deja}, in our training scenario, since our sampling operation is performed per batch rather than per token, the cost of extracting rows and columns from the matrices is relatively small, and kernel fusion is not necessary.

\subsection{Implementation of Index Generator}
The index generator simply generates random numbers as sampling indices. However, since it operates on the CPU, and once the indices are generated, they need to be transferred to the GPU memory. Executing it as part of the model before each forward pass could result in unnecessary time overhead. To optimize the training process as much as possible, we use an additional thread to run the index generator asynchronously to the model training. Once the index generator generates the next set of indices, it places them in a queue. When the model needs to sample, it retrieves the values from the queue. This completely eliminates the overhead of the  index generator.


%% file: example_paper.bbl
\begin{thebibliography}{34}
\providecommand{\natexlab}[1]{#1}
\providecommand{\url}[1]{\texttt{#1}}
\expandafter\ifx\csname urlstyle\endcsname\relax
  \providecommand{\doi}[1]{doi: #1}\else
  \providecommand{\doi}{doi: \begingroup \urlstyle{rm}\Url}\fi

\bibitem[Anand et~al.(2023)Anand, Nussbaum, Duderstadt, Schmidt, and Mulyar]{gpt4all}
Anand, Y., Nussbaum, Z., Duderstadt, B., Schmidt, B., and Mulyar, A.
\newblock Gpt4all: Training an assistant-style chatbot with large scale data distillation from gpt-3.5-turbo.
\newblock \url{https://github.com/nomic-ai/gpt4all}, 2023.

\bibitem[Ba et~al.(2016)Ba, Kiros, and Hinton]{layernorm}
Ba, J.~L., Kiros, J.~R., and Hinton, G.~E.
\newblock Layer normalization.
\newblock \emph{arXiv preprint arXiv:1607.06450}, 2016.

\bibitem[Brown et~al.(2020)Brown, Mann, Ryder, Subbiah, Kaplan, Dhariwal, Neelakantan, Shyam, Sastry, Askell, et~al.]{gpt3}
Brown, T., Mann, B., Ryder, N., Subbiah, M., Kaplan, J.~D., Dhariwal, P., Neelakantan, A., Shyam, P., Sastry, G., Askell, A., et~al.
\newblock Language models are few-shot learners.
\newblock \emph{Advances in neural information processing systems}, 33:\penalty0 1877--1901, 2020.

\bibitem[Chen et~al.(2022)Chen, Yin, Shang, Jiang, Qin, Wang, Wang, Chen, Liu, and Liu]{bert2bert}
Chen, C., Yin, Y., Shang, L., Jiang, X., Qin, Y., Wang, F., Wang, Z., Chen, X., Liu, Z., and Liu, Q.
\newblock bert2bert: Towards reusable pretrained language models.
\newblock In \emph{Proceedings of the 60th Annual Meeting of the Association for Computational Linguistics (Volume 1: Long Papers)}, pp.\  2134--2148, 2022.

\bibitem[Chen et~al.(2016)Chen, Goodfellow, and Shlens]{net2net}
Chen, T., Goodfellow, I.~J., and Shlens, J.
\newblock Net2net: Accelerating learning via knowledge transfer.
\newblock In Bengio, Y. and LeCun, Y. (eds.), \emph{4th International Conference on Learning Representations, {ICLR} 2016, San Juan, Puerto Rico, May 2-4, 2016, Conference Track Proceedings}, 2016.
\newblock URL \url{http://arxiv.org/abs/1511.05641}.

\bibitem[Chen et~al.(2021)Chen, Cheng, Wang, Gan, Wang, and Liu]{earlybert}
Chen, X., Cheng, Y., Wang, S., Gan, Z., Wang, Z., and Liu, J.
\newblock Early{\{}bert{\}}: Efficient {\{}bert{\}} training via early-bird lottery tickets, 2021.
\newblock URL \url{https://openreview.net/forum?id=I-VfjSBzi36}.

\bibitem[Choromanski et~al.(2021)Choromanski, Likhosherstov, Dohan, Song, Gane, Sarlos, Hawkins, Davis, Mohiuddin, Kaiser, Belanger, Colwell, and Weller]{performer}
Choromanski, K.~M., Likhosherstov, V., Dohan, D., Song, X., Gane, A., Sarlos, T., Hawkins, P., Davis, J.~Q., Mohiuddin, A., Kaiser, L., Belanger, D.~B., Colwell, L.~J., and Weller, A.
\newblock Rethinking attention with performers.
\newblock In \emph{International Conference on Learning Representations}, 2021.
\newblock URL \url{https://openreview.net/forum?id=Ua6zuk0WRH}.

\bibitem[Dao et~al.(2021)Dao, Chen, Liang, Yang, Song, Rudra, and Re]{dao2021pixelated}
Dao, T., Chen, B., Liang, K., Yang, J., Song, Z., Rudra, A., and Re, C.
\newblock Pixelated butterfly: Simple and efficient sparse training for neural network models.
\newblock \emph{arXiv preprint arXiv:2112.00029}, 2021.

\bibitem[Dao et~al.(2022{\natexlab{a}})Dao, Chen, Sohoni, Desai, Poli, Grogan, Liu, Rao, Rudra, and R{\'e}]{dao2022monarch}
Dao, T., Chen, B., Sohoni, N.~S., Desai, A., Poli, M., Grogan, J., Liu, A., Rao, A., Rudra, A., and R{\'e}, C.
\newblock Monarch: Expressive structured matrices for efficient and accurate training.
\newblock In \emph{International Conference on Machine Learning}, pp.\  4690--4721. PMLR, 2022{\natexlab{a}}.

\bibitem[Dao et~al.(2022{\natexlab{b}})Dao, Fu, Ermon, Rudra, and R{\'e}]{flashattention}
Dao, T., Fu, D., Ermon, S., Rudra, A., and R{\'e}, C.
\newblock Flashattention: Fast and memory-efficient exact attention with io-awareness.
\newblock \emph{Advances in Neural Information Processing Systems}, 35:\penalty0 16344--16359, 2022{\natexlab{b}}.

\bibitem[Du et~al.(2022)Du, Huang, Dai, Tong, Lepikhin, Xu, Krikun, Zhou, Yu, Firat, et~al.]{glam}
Du, N., Huang, Y., Dai, A.~M., Tong, S., Lepikhin, D., Xu, Y., Krikun, M., Zhou, Y., Yu, A.~W., Firat, O., et~al.
\newblock Glam: Efficient scaling of language models with mixture-of-experts.
\newblock In \emph{International Conference on Machine Learning}, pp.\  5547--5569. PMLR, 2022.

\bibitem[Fedus et~al.(2022)Fedus, Zoph, and Shazeer]{switch_transformer}
Fedus, W., Zoph, B., and Shazeer, N.
\newblock Switch transformers: Scaling to trillion parameter models with simple and efficient sparsity.
\newblock \emph{The Journal of Machine Learning Research}, 23\penalty0 (1):\penalty0 5232--5270, 2022.

\bibitem[Frankle \& Carbin(2019)Frankle and Carbin]{lottery}
Frankle, J. and Carbin, M.
\newblock The lottery ticket hypothesis: Finding sparse, trainable neural networks.
\newblock In \emph{International Conference on Learning Representations}, 2019.
\newblock URL \url{https://openreview.net/forum?id=rJl-b3RcF7}.

\bibitem[Gu \& Dao(2023)Gu and Dao]{gu2023mamba}
Gu, A. and Dao, T.
\newblock Mamba: Linear-time sequence modeling with selective state spaces, 2023.

\bibitem[Huang et~al.(2016)Huang, Sun, Liu, Sedra, and Weinberger]{stochastic_depth}
Huang, G., Sun, Y., Liu, Z., Sedra, D., and Weinberger, K.
\newblock Deep networks with stochastic depth.
\newblock volume 9908, pp.\  646--661, 10 2016.
\newblock ISBN 978-3-319-46492-3.
\newblock \doi{10.1007/978-3-319-46493-0_39}.

\bibitem[Kitaev et~al.(2020)Kitaev, Kaiser, and Levskaya]{Reformer}
Kitaev, N., Kaiser, L., and Levskaya, A.
\newblock Reformer: The efficient transformer.
\newblock In \emph{International Conference on Learning Representations}, 2020.
\newblock URL \url{https://openreview.net/forum?id=rkgNKkHtvB}.

\bibitem[Li et~al.(2023{\natexlab{a}})Li, Allal, Zi, Muennighoff, Kocetkov, Mou, Marone, Akiki, Li, Chim, Liu, Zheltonozhskii, Zhuo, Wang, Dehaene, Davaadorj, Lamy-Poirier, Monteiro, Shliazhko, Gontier, Meade, Zebaze, Yee, Umapathi, Zhu, Lipkin, Oblokulov, Wang, Murthy, Stillerman, Patel, Abulkhanov, Zocca, Dey, Zhang, Fahmy, Bhattacharyya, Yu, Singh, Luccioni, Villegas, Kunakov, Zhdanov, Romero, Lee, Timor, Ding, Schlesinger, Schoelkopf, Ebert, Dao, Mishra, Gu, Robinson, Anderson, Dolan-Gavitt, Contractor, Reddy, Fried, Bahdanau, Jernite, Ferrandis, Hughes, Wolf, Guha, von Werra, and de~Vries]{li2023starcoder}
Li, R., Allal, L.~B., Zi, Y., Muennighoff, N., Kocetkov, D., Mou, C., Marone, M., Akiki, C., Li, J., Chim, J., Liu, Q., Zheltonozhskii, E., Zhuo, T.~Y., Wang, T., Dehaene, O., Davaadorj, M., Lamy-Poirier, J., Monteiro, J., Shliazhko, O., Gontier, N., Meade, N., Zebaze, A., Yee, M.-H., Umapathi, L.~K., Zhu, J., Lipkin, B., Oblokulov, M., Wang, Z., Murthy, R., Stillerman, J., Patel, S.~S., Abulkhanov, D., Zocca, M., Dey, M., Zhang, Z., Fahmy, N., Bhattacharyya, U., Yu, W., Singh, S., Luccioni, S., Villegas, P., Kunakov, M., Zhdanov, F., Romero, M., Lee, T., Timor, N., Ding, J., Schlesinger, C., Schoelkopf, H., Ebert, J., Dao, T., Mishra, M., Gu, A., Robinson, J., Anderson, C.~J., Dolan-Gavitt, B., Contractor, D., Reddy, S., Fried, D., Bahdanau, D., Jernite, Y., Ferrandis, C.~M., Hughes, S., Wolf, T., Guha, A., von Werra, L., and de~Vries, H.
\newblock Starcoder: may the source be with you!
\newblock 2023{\natexlab{a}}.

\bibitem[Li et~al.(2020)Li, Wallace, Shen, Lin, Keutzer, Klein, and Gonzalez]{over_para}
Li, Z., Wallace, E., Shen, S., Lin, K., Keutzer, K., Klein, D., and Gonzalez, J.
\newblock Train big, then compress: Rethinking model size for efficient training and inference of transformers.
\newblock In \emph{International Conference on machine learning}, pp.\  5958--5968. PMLR, 2020.

\bibitem[Li et~al.(2023{\natexlab{b}})Li, You, Bhojanapalli, Li, Rawat, Reddi, Ye, Chern, Yu, Guo, et~al.]{conditional_sparsity}
Li, Z., You, C., Bhojanapalli, S., Li, D., Rawat, A.~S., Reddi, S., Ye, K., Chern, F., Yu, F., Guo, R., et~al.
\newblock The lazy neuron phenomenon: On emergence of activation sparsity in transformers.
\newblock In \emph{Conference on Parsimony and Learning (Recent Spotlight Track)}, 2023{\natexlab{b}}.

\bibitem[Liu et~al.(2023{\natexlab{a}})Liu, Xie, Li, and Ma]{liu2023same}
Liu, H., Xie, S.~M., Li, Z., and Ma, T.
\newblock Same pre-training loss, better downstream: Implicit bias matters for language models.
\newblock In \emph{International Conference on Machine Learning}, pp.\  22188--22214. PMLR, 2023{\natexlab{a}}.

\bibitem[Liu et~al.(2023{\natexlab{b}})Liu, Wang, Dao, Zhou, Yuan, Song, Shrivastava, Zhang, Tian, Re, et~al.]{liu2023deja}
Liu, Z., Wang, J., Dao, T., Zhou, T., Yuan, B., Song, Z., Shrivastava, A., Zhang, C., Tian, Y., Re, C., et~al.
\newblock Deja vu: Contextual sparsity for efficient llms at inference time.
\newblock In \emph{International Conference on Machine Learning}, pp.\  22137--22176. PMLR, 2023{\natexlab{b}}.

\bibitem[Loshchilov \& Hutter(2019)Loshchilov and Hutter]{adamw}
Loshchilov, I. and Hutter, F.
\newblock Decoupled weight decay regularization.
\newblock In \emph{International Conference on Learning Representations}, 2019.
\newblock URL \url{https://openreview.net/forum?id=Bkg6RiCqY7}.

\bibitem[Ma et~al.(2024)Ma, Chen, Wang, Xu, Li, Sun, Zhu, Fan, and Yu]{mada2024sparsityaccelerated}
Ma, D., Chen, L., Wang, P., Xu, H., Li, H., Sun, L., Zhu, S., Fan, S., and Yu, K.
\newblock Sparsity-accelerated training for large language models.
\newblock In \emph{The 62nd Annual Meeting of the Association for Computational Linguistics}, 2024.
\newblock URL \url{https://openreview.net/forum?id=HvofKj7jlC}.

\bibitem[Pal et~al.(2020)Pal, Lane, Vidal, and Haeffele]{structural_dropout}
Pal, A., Lane, C., Vidal, R., and Haeffele, B.~D.
\newblock On the regularization properties of structured dropout.
\newblock In \emph{Proceedings of the IEEE/CVF conference on computer vision and pattern recognition}, pp.\  7671--7679, 2020.

\bibitem[Paperno et~al.(2016)Paperno, Kruszewski, Lazaridou, Pham, Bernardi, Pezzelle, Baroni, Boleda, and Fern{\'a}ndez]{lambada}
Paperno, D., Kruszewski, G., Lazaridou, A., Pham, N.~Q., Bernardi, R., Pezzelle, S., Baroni, M., Boleda, G., and Fern{\'a}ndez, R.
\newblock The {LAMBADA} dataset: Word prediction requiring a broad discourse context.
\newblock In Erk, K. and Smith, N.~A. (eds.), \emph{Proceedings of the 54th Annual Meeting of the Association for Computational Linguistics (Volume 1: Long Papers)}, pp.\  1525--1534, Berlin, Germany, August 2016. Association for Computational Linguistics.
\newblock \doi{10.18653/v1/P16-1144}.
\newblock URL \url{https://aclanthology.org/P16-1144}.

\bibitem[Radford et~al.(2019)Radford, Wu, Child, Luan, Amodei, Sutskever, et~al.]{gpt2}
Radford, A., Wu, J., Child, R., Luan, D., Amodei, D., Sutskever, I., et~al.
\newblock Language models are unsupervised multitask learners.
\newblock \emph{OpenAI blog}, 1\penalty0 (8):\penalty0 9, 2019.

\bibitem[Rajpurkar et~al.(2016)Rajpurkar, Zhang, Lopyrev, and Liang]{squad}
Rajpurkar, P., Zhang, J., Lopyrev, K., and Liang, P.
\newblock {SQ}u{AD}: 100,000+ questions for machine comprehension of text.
\newblock In Su, J., Duh, K., and Carreras, X. (eds.), \emph{Proceedings of the 2016 Conference on Empirical Methods in Natural Language Processing}, pp.\  2383--2392, Austin, Texas, November 2016. Association for Computational Linguistics.
\newblock \doi{10.18653/v1/D16-1264}.
\newblock URL \url{https://aclanthology.org/D16-1264}.

\bibitem[Schwartz et~al.(2020)Schwartz, Dodge, Smith, and Etzioni]{schwartz2020green}
Schwartz, R., Dodge, J., Smith, N.~A., and Etzioni, O.
\newblock Green ai.
\newblock \emph{Communications of the ACM}, 63\penalty0 (12):\penalty0 54--63, 2020.

\bibitem[Shen et~al.(2022)Shen, Walsh, Keutzer, Dodge, Peters, and Beltagy]{shen2022staged}
Shen, S., Walsh, P., Keutzer, K., Dodge, J., Peters, M., and Beltagy, I.
\newblock Staged training for transformer language models.
\newblock In \emph{International Conference on Machine Learning}, pp.\  19893--19908. PMLR, 2022.

\bibitem[Soboleva et~al.(2023)Soboleva, Al-Khateeb, Myers, Steeves, Hestness, and Dey]{cerebras2023slimpajama}
Soboleva, D., Al-Khateeb, F., Myers, R., Steeves, J.~R., Hestness, J., and Dey, N.
\newblock {SlimPajama: A 627B token cleaned and deduplicated version of RedPajama}, 2023.
\newblock URL \url{https://huggingface.co/datasets/cerebras/SlimPajama-627B}.

\bibitem[Vaswani et~al.(2017)Vaswani, Shazeer, Parmar, Uszkoreit, Jones, Gomez, Kaiser, and Polosukhin]{vaswani2017attention}
Vaswani, A., Shazeer, N., Parmar, N., Uszkoreit, J., Jones, L., Gomez, A.~N., Kaiser, {\L}., and Polosukhin, I.
\newblock Attention is all you need.
\newblock \emph{Advances in neural information processing systems}, 30, 2017.

\bibitem[Wang et~al.(2018)Wang, Singh, Michael, Hill, Levy, and Bowman]{glue}
Wang, A., Singh, A., Michael, J., Hill, F., Levy, O., and Bowman, S.~R.
\newblock Glue: A multi-task benchmark and analysis platform for natural language understanding.
\newblock \emph{arXiv preprint arXiv:1804.07461}, 2018.

\bibitem[Yao et~al.(2023)Yao, Zhang, Li, and Wang]{MSG}
Yao, Y., Zhang, Z., Li, J., and Wang, Y.
\newblock 2x faster language model pre-training via masked structural growth.
\newblock \emph{arXiv preprint arXiv:2305.02869}, 2023.

\bibitem[Zhang et~al.(2024)Zhang, Zeng, Wang, and Lu]{tinyllama}
Zhang, P., Zeng, G., Wang, T., and Lu, W.
\newblock Tinyllama: An open-source small language model, 2024.

\end{thebibliography}
